\documentclass[letterpaper, 10 pt, journal, twoside]{IEEEtran}

\usepackage{adjustbox}
\usepackage{algorithm,algorithmic}
\usepackage{amsfonts, amssymb}
\usepackage{cite}
\usepackage{color}
\usepackage{flushend}
\usepackage{graphicx}
\usepackage{multirow}
\usepackage[labelformat=simple]{subcaption}

\usepackage{tabularx}
\usepackage{textcomp}
\usepackage{url}
\usepackage{verbatim}
\usepackage{wrapfig}
\newcommand{\basicrrt}{Basic RRT}
\newcommand{\cfree}{\mathcal{C}_{free}}
\newcommand{\cspace}{\mathcal{C}} 
\newcommand{\cobst}{\mathcal{C}_{obst}}

\newcommand{\eet}{{\sc EET}}
\newcommand{\hasap}{Hierarchical Annotated-Skeleton guided PRM}
\newcommand{\hasrrtlong}{Hierarchical Annotated-Skeleton Guided RRT}
\newcommand{\hasrrt}{HAS-RRT}
\newcommand{\drrrt}{DR-RRT}
\newcommand{\drbrrt}{Dynamic Region-biased RRT}
\newcommand{\drprm}{Dynamic Region Sampling with PRM}

\newcommand{\rrt}{RRT}

\newcommand{\qskeleton}{Query Skeleton}
\newcommand{\qrand}{q_{rand}}
\newcommand{\qnear}{q_{near}}

\newcommand{\qstart}{q_{start}}
\newcommand{\qgoal}{q_{goal}}

\begin{document}

\title{HAS-RRT: RRT-based Motion Planning using Topological Guidance} 

\author{Diane Uwacu$^{1*}$, Ananya Yammanuru$^{2*}$, Keerthana Nallamotu$^{2}$, Vasu Chalasani$^{2}$,\\Marco Morales$^{2,3}$, Nancy M. Amato$^{2}$%
 \thanks{Manuscript received: September 15, 2024; Revised December 25, 2024; Accepted March 25, 2025.}
 \thanks{This paper was recommended for publication by
Editor Aniket Bera upon evaluation of the Associate Editor and Reviewers’
comments. This work was supported in part by the U.S. National Science Foundation's ``Expeditions: Mind in Vitro: Computing with Living Neurons"  under award No. IIS-2123781, and by the IBM-Illinois Discovery Accelerator Institute and the Center for Networked Intelligent Components and Environments (C-NICE) at the University of Illinois.
Yammanuru was supported in part by an NSF GRFP. Morales was supported in part by Asociación Mexicana de Cultura A.C.}
 \thanks{$^*$ Equal contribution.}
 \thanks{$^{1}$Department of Computer Science at Mt. Holyoke College, South Hadley, MA, USA
        {\tt\small duwacu@    mtholyoke.edu}}%
\thanks{$^{2}$ Department of Computer Science at the University of Illinois at Urbana-Champaign, Urbana, IL, USA
        {\tt\small (ananyay2, kn19, sc83, moralesa, namato)@illinois.edu}}%
\thanks{$^{3}$ Department of Computer Science at Instituto Tecnológico Autónomo de México (ITAM), Mexico City, México.}%
}

\markboth{IEEE Robotics and Automation Letters. Preprint Version. April, 2025}
{Uwacu \MakeLowercase{\textit{et al.}}: HAS-RRT: RRT-based Motion Planning using Topological Guidance}

\maketitle
\begin{abstract}
    We present a hierarchical RRT-based motion planning strategy, \hasrrtlong\ (\hasrrt), guided by a workspace skeleton, to solve motion planning problems.
    \hasrrt\ provides up to a 91\% runtime reduction and builds a tree at least 30\% smaller than competitors while still finding competitive-cost paths. 
    This is because our strategy prioritizes paths indicated by the workspace guidance to efficiently find a valid motion plan for the robot. 
    Existing methods either rely too heavily on workspace guidance or have difficulty finding narrow passages. 
    By taking advantage of the assumptions that the workspace skeleton provides, \hasrrt\ is able to build a smaller tree and find a path faster than its competitors. 
    Additionally, we show that \hasrrt\ is robust to the quality of workspace guidance provided and that, in a worst-case scenario where the workspace skeleton provides no additional insight, our method performs comparably to an unguided method. 
\end{abstract}

\section{Introduction}
\label{sec:introduction}

Motion planning algorithms have applications in many fields, from robotics \cite{englot2012AAAI} to microbiology \cite{baghvsl-urfpdl}. 
Exact motion planning is PSPACE-hard and
intractable, as deterministic algorithms are exponential in the number of degrees of freedom of the robot ~\cite{r-cpmpg-79}.
We can use sampling-based planners to approximate the robot's high dimensional planning space (the configuration space, or $\cspace$) \cite{lk-rrtpp-00, kslo-prpp-96}. 

\begin{figure}[t]
  \centering
\begin{subfigure}{.2\textwidth}
  \centering
    \includegraphics[width=1\textwidth]{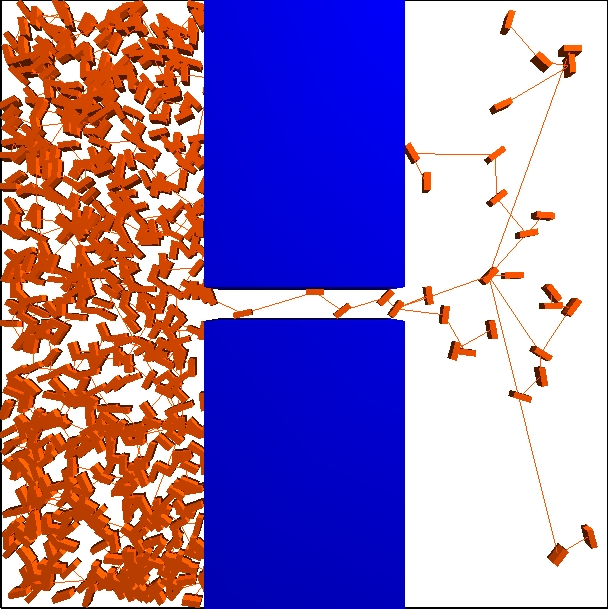}
  \caption{Basic RRT \cite{lk-rrtpp-00}}
  \label{fig:tree-rrt}
\end{subfigure}
  \centering
\begin{subfigure}{.2\textwidth}
  \centering
    \includegraphics[width=1\textwidth]{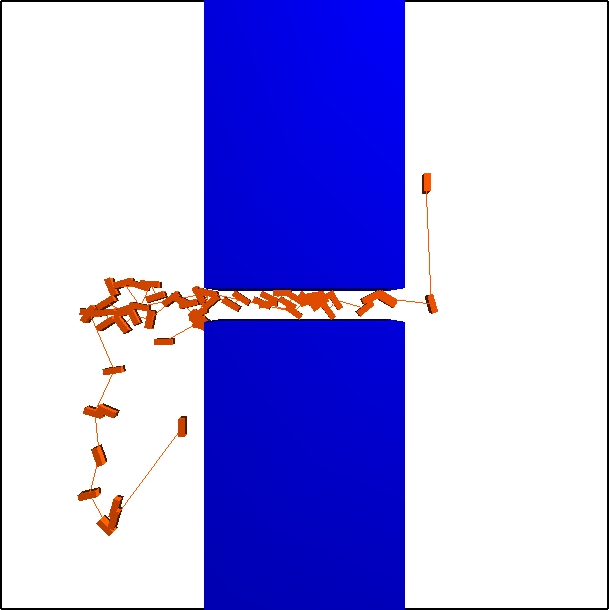}
  \caption{DR-RRT \cite{dsba-drbrrt-16}}
  \label{fig:tree-drrrt}
\end{subfigure}
  \centering
\begin{subfigure}{.2\textwidth}
  \centering
    \includegraphics[width=1\textwidth]{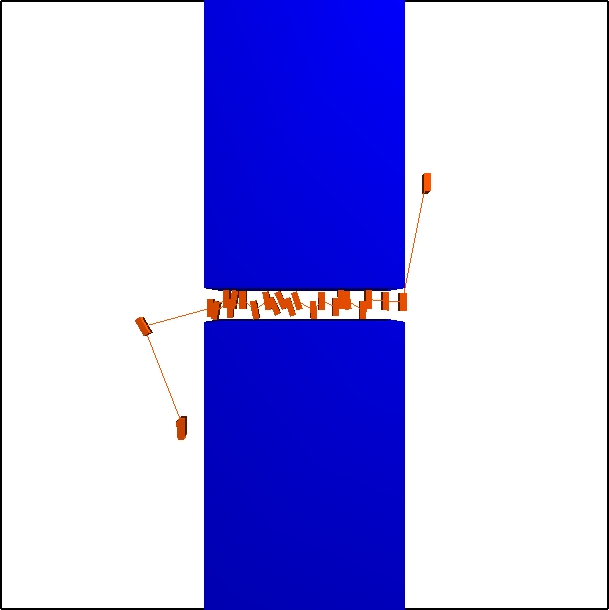}
  \caption{EET \cite{Rickert_balancingexploration}}
  \label{fig:tree-eet}
\end{subfigure}
  \centering
\begin{subfigure}{.2\textwidth}
  \centering
    \includegraphics[width=1\textwidth]{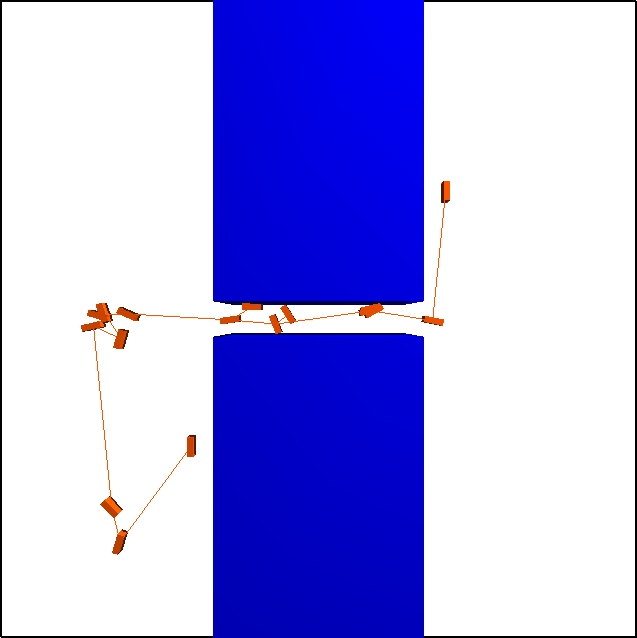}
  \caption{HAS-RRT}
  \label{fig:tree-hasrrt}
\end{subfigure}
  \caption{
    Trees from \rrt~variants in a narrow passage problem. \hasrrt~ shows a better balance of exploring the open space and narrow passage.}
  \label{fig:roadmaps}
\end{figure}

\begin{figure}[]
  \centering
\begin{subfigure}{.2\textwidth}
  \centering
    \includegraphics[width=1\textwidth]{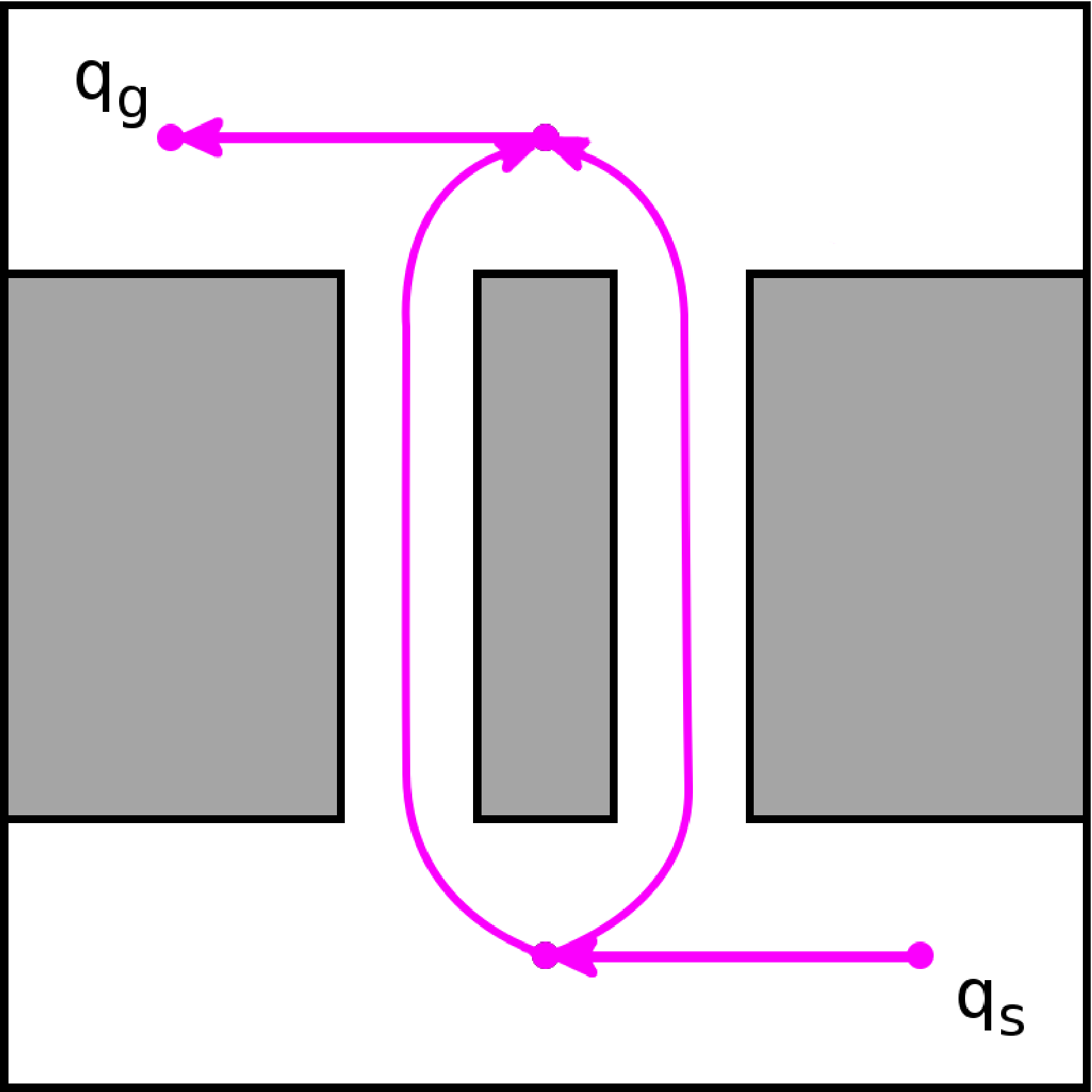}
  \caption{}
  \label{fig:dbrrt-step1}
\end{subfigure}
  \centering
\begin{subfigure}{.2\textwidth}
  \centering
    \includegraphics[width=1\textwidth]{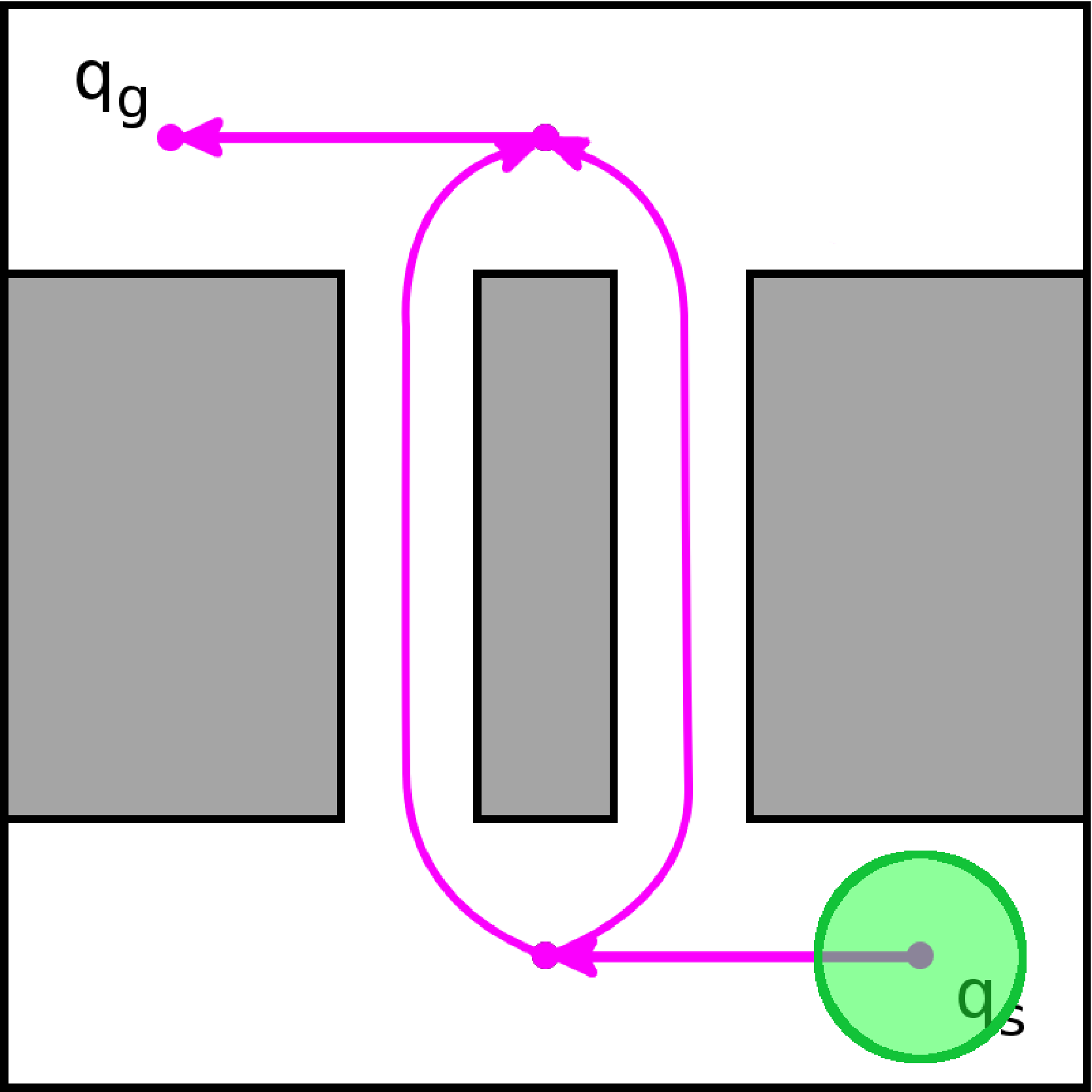}
  \caption{}
  \label{fig:dbrrt-step2}
\end{subfigure}
  \centering
\begin{subfigure}{.2\textwidth}
  \centering
    \includegraphics[width=1\textwidth]{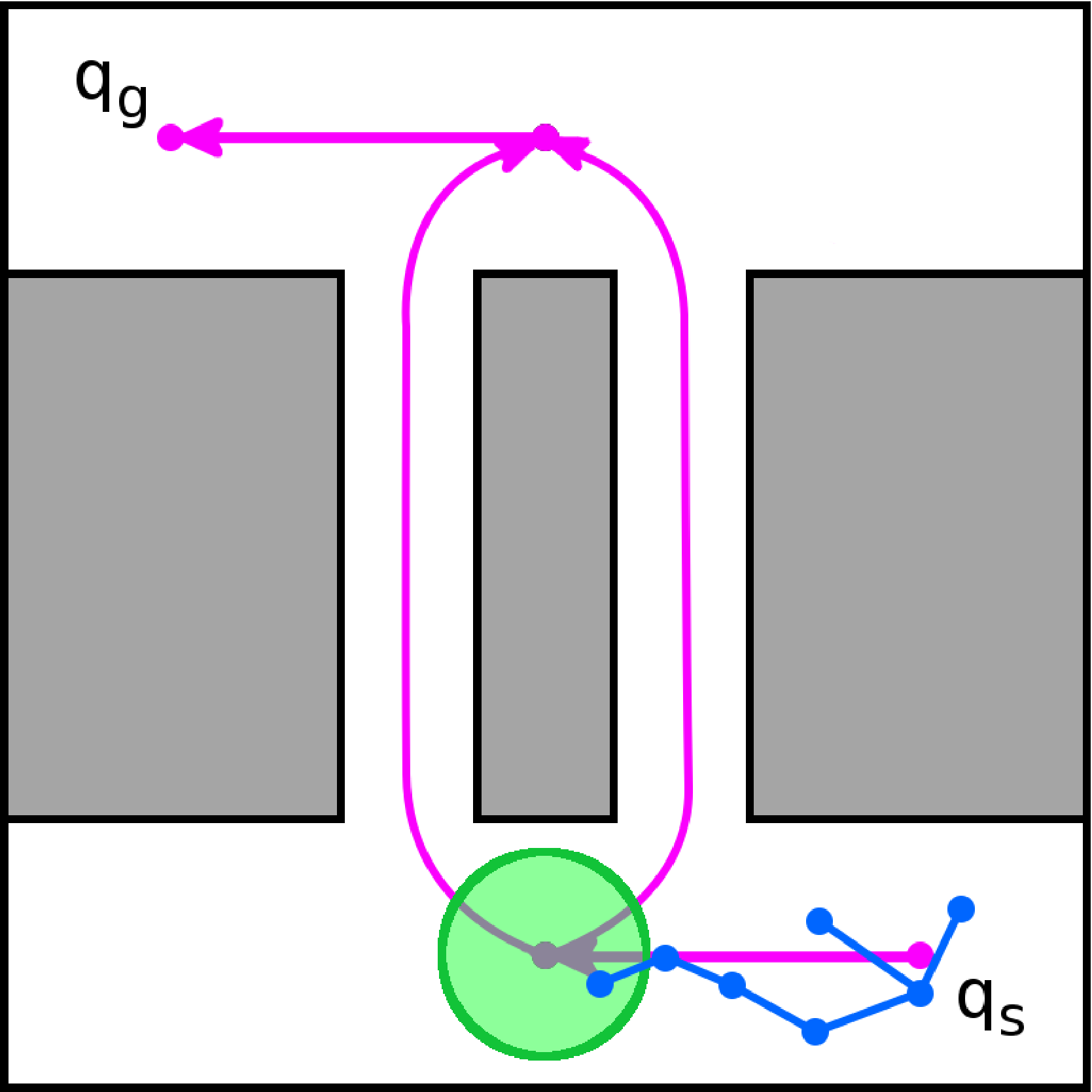}
  \caption{}
  \label{fig:dbrrt-step3}
\end{subfigure}
  \centering
\begin{subfigure}{.2\textwidth}
  \centering
    \includegraphics[width=1\textwidth]{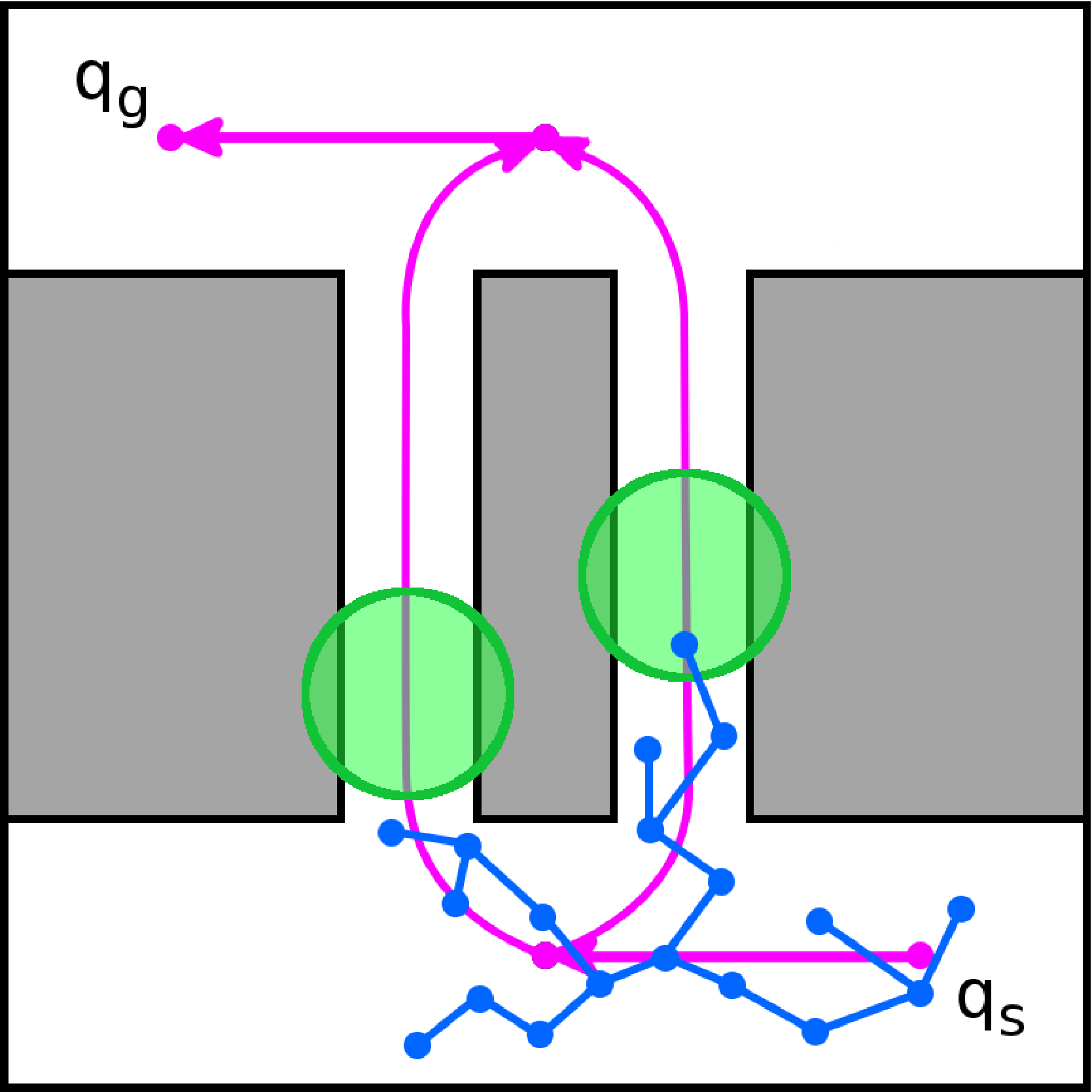}
  \caption{}
  \label{fig:dbrrt-step4}
\end{subfigure}
  \caption{
    Example execution of \drbrrt:
    (a) \qskeleton\ (magenta) computed from a query $\{q_s, q_g\}$;
    (b) initial region (green) placed at the source vertex of the \qskeleton;
    (c) region-biased \rrt\ growth (blue); and
    (d) multiple active regions (green) guiding the tree (blue) among multiple
    embedded arcs of the \qskeleton\ (magenta) is shown.
  }
  \label{fig:dynamicregion:example}
\end{figure}

Narrow passages in $\cspace$ are a challenge for sampling-based algorithms. The probability of randomly discovering those narrow passages is very low, hence the necessity to find ways to 
provide guidance to the planning process. 
Initial methods including OBPRM \cite{abdjv-obprm-98}, OBRRT \cite{rtla-obrrt-06}, 
and MAPRM \cite{was-maprm-99} attempt to address the narrow passage problem by biasing the sampling
process using an approximation of the obstacle space (the portion of $\cspace$ occupied by 
obstacles) as guidance. However, these methods can be slow in dense or cluttered environments because 
they require lots of calls to a collision checker.
We can identify narrow passages with the help of information about the connectivity of the workspace. 
Such information can be encoded in a graph called a workspace skeleton and used to guide the exploration of the configuration space.
The Dynamic Region-biased RRT \cite{dsba-drbrrt-16} algorithm is a skeleton-based sampling strategy that showed that the workspace skeleton can be a reliable guide. 
However, it does not take full advantage of the guidance that a workspace skeleton can provide. \drrrt\ takes small steps along the workspace skeleton, which can be slow and unnecessary (see Figure \ref{fig:dynamicregion:example})
. 

This work presents the Hierarchical Annotated Skeleton RRT (\hasrrt) method to quickly find workspace-guided paths in a single-query environment. Similar to \cite{dsba-drbrrt-16}, \hasrrt~ uses the workspace skeleton to search for paths. However, \hasrrt~ initially relies on the connectivity of the workspace to make large steps in its exploration and only retracts to focused local exploration when necessary. 
This approach makes \hasrrt~more robust to poor-quality skeletons than previous methods.
such as environment clearance or priority through certain passages
, the strategy can be adapted to fit different motion planning applications.
When the workspace skeleton contains workspace paths that can be extended to $\cspace$ paths, \hasrrt\ exploits the guidance fully, and when the skeleton is not perfectly indicative of the motion planning solution, the method can use directional hints from such a skeleton and find a solution with additional exploration more gracefully than the other methods.
This work most applies to robots whose workspace and configuration space are closely related, such as mobile robots. 

Our results show how the \hasrrt\ method achieves improved run time, efficiency, and scalability compared to the basic RRT strategy \cite{lk-rrtpp-00}, the Exploration-Exploitation tree (\eet) strategy \cite{Rickert_balancingexploration}, and the Dynamic Region RRT (\drrrt) strategy \cite{dsba-drbrrt-16}. 
Our runtime improves by at least 59\% and as much as 91\% on average compared to \drrrt\ in each of our test environments.
In addition, as shown in Figure \ref{fig:roadmaps}, \hasrrt\ yields a sparser tree with exploration focused on critical regions of $\cspace$ compared to those strategies.


Specifically, our contributions include:
\begin{itemize}
    \item A novel workspace guidance algorithm, \hasrrtlong, which can rapidly discover paths indicated by a workspace skeleton. 
    \item An empirical validation  that demonstrates \hasrrt's reliable efficiency in run time and path quality across various environments and highlights it as the best strategy for workspace-guided planning.
    \item 
    A study that shows how \hasrrt\ reacts to varying levels of guidance and shows that with extremely poor guidance, \hasrrt's behavior is similar to that of \rrt. 
    Previous work using workspace skeletons does not include an analysis of performance under varying qualities of guidance.
\end{itemize}
\section{Preliminaries and Previous Work}
\label{sec:prev-work}
Section \ref{sec:mp-prelims} provides an overview of the motion planning problem and some basic algorithms used to solve it. Section \ref{sec:bkgd-workspace} covers different types of workspace guidance, including the workspace skeleton, and relevant previous methods which use workspace guidance. 

\subsection{Motion Planning Preliminaries and Sampling-based Planning}
\label{sec:mp-prelims}
A robot's pose is described by its configuration, and the set of all possible configurations within an environment is called the \textit{configuration space}, or $\cspace$. The dimension of $\cspace$ is the same as the number of degrees of freedom of the robot. $\cspace$ is partitioned into $\cfree$ (the set of valid configurations, or the free space) and $\cobst$ (the set of invalid configurations, or the obstacle space).

Given a start configuration $\qstart$ and a goal configuration $\qgoal$, the goal of a motion planning problem is to find a path from $\qstart$ to $\qgoal$ which lies entirely in $\cfree$. 
In other words, the goal is to find a path  $p:[0, 1]\rightarrow \cspace$, where $p(0) = \qstart$ and $p(1) = \qgoal$. 
This is called a \textit{query}.

Sampling-based planning algorithms approximate $\cspace$, since it is difficult to calculate exactly. These methods randomly generate configurations in $\cspace$ and connect them to build a representational graph. Examples of such methods include probabilistic roadmaps (PRM) \cite{kslo-prpp-96} , which constructs a graph, and rapidly exploring random trees (RRT) \cite{lk-rrtpp-00}, which constructs a tree.  
Graph-based methods can be reused for multiple queries, while tree-based methods must be used for one query at a time.
In this work, we focus on RRT and some of its variations to study problems related to accessibility. 
 
RRT algorithms are generally used to solve single-query motion planning problems. The tree starts at the start configuration, $\qstart$. In each algorithm iteration, a random direction, $\qrand$, is sampled, and its nearest neighbor $\qnear$ in the tree is identified. An extension from $\qnear$ to $\qrand$ is attempted to expand the tree. After enough iterations, the tree reaches the goal, and the query is solved.

Since the direction to expand the tree is randomly selected, the basic RRT algorithm is probabilistically complete. This means that that if a solution exists, the probability of finding it increases as more time and resources are spent growing the tree \cite{lk-rrtpp-00}. This, however, also implies that the algorithm struggles in more constrained environments. For example, if an environment has many dead ends, the basic RRT algorithm may waste time exploring unnecessary directions.

\subsection{Workspace-Guided Planning}
\label{sec:bkgd-workspace}

Information about the workspace can be used to bias sampling towards $\cfree$ for more optimal paths or faster search time. Many methods use the workspace to guide planning, and here we describe a few.

Some methods perform a preliminary search of the workspace before sampling. The Exploration-Exploitation Trees (\eet) algorithm \cite{Rickert_balancingexploration} first explores the workspace by randomly growing a tree of spheres from the start position to the goal. This is called the `Wavefront' expansion tree, and with it, \eet~grows an \rrt~in $\cspace$, exploiting the pre-explored sphere regions. If exploitation fails in difficult regions, the planner gradually shifts its behavior to exploration with regular sampling-based planning. 
The wavefront exploration process is highly sensitive to the positioning of the start and goal configurations and the variability of region sizes in the environment. 
In addition, the information about the workspace is acquired by using clearance-based spheres that are randomly expanded to cover the free workspace. In environments with narrow passages, this process often takes too long.

Workspace Decomposition Strategies \cite{kh-wisprp-04, bo-uwignsprp-04, kh-wco-06} steer sampling in the free workspace. For example, SyCLoP \cite{pkv-mpdsclp-10} uses an RRT to sample frontier decomposition regions. The decomposition of the workspace is limited as a guide, however, because it does not contain information about the topology of the workspace. 

Some methods attempt to use the medial axis, the set of all points equidistant to at least 2 obstacles, to help guide sampling. MAPRM and aMAPRM \cite{hk-fuwmapp-00, yb-asdppbama-04} generate all their PRM samples along the medial axis. MARRT \cite{dgta-marrt-14} attempts to do the same, but with an RRT tree expansion.

A workspace skeleton has been shown to provide better guidance for sampling. A workspace skeleton is a graph that denotes the connectivity of the workspace \cite{blk-tcsrp-2012}. The vertices of the skeleton denote regions in the workspace, and the edges denote connectivity between them. A skeleton can be computed geometrically or provided by a user. Examples of workspace skeletons can be found in Figure \ref{fig:envs}

Workspace skeletons are more effective when $\cspace$ is a subset of the workspace. When sampling along a workspace skeleton, a sample directly corresponds to a point in $\cspace$. When $\cspace$ does not correspond to the workspace (e.g. manipulator robots, where $\cspace$ has much different topology from the workspace), it can be difficult to find a transformation from a sample taken in the workspace to a point in the free configuration space. EET \cite{Rickert_balancingexploration} addresses this by corresponding a point in the the workspace to the location of the end effector, however, this approach does not guarantee finding a path if one exists. 

A precursor of the work presented in this paper is the \drbrrt \cite{dsba-drbrrt-16}, detailed in Figure \ref{fig:dynamicregion:example}. The workspace has obstacles and a Reeb graph \cite{dn-eacrg-09} skeleton. 
To speed up the process, the workspace skeleton has been pruned and directed to only have parts connecting the current start to the current goal.
A sampling region is instantiated close to $\qstart$ and expanded along the nearest skeleton edge until it reaches the next skeleton vertex and splits into several regions, one for each adjacent edge. 
The process ends when one of the regions is close enough to the goal to solve the query. 

\drbrrt~ is efficient because the tree is guided through the connected free space, simplifying the narrow passage problem and preventing the tree from being stagnant.
However, \drbrrt~ does unnecessary exploration by expanding the dynamic region in small steps along the skeleton edge.

Other similar methods which use a workspace skeleton include \drprm \cite{suda-tgrcdrs-20} and \hasap \cite{uyma-hpasg-22}, both of which use the workspace skeleton as guidance for building and querying a probabilistic roadmap. 
While RRT-based methods such as \drbrrt~prune and direct the portions of the workspace skeleton related to a query, these PRM-based methods use the whole skeleton to build a roadmap. 



Mathematically, we
define the workspace skeleton as a graph $G_s = (V_s, E_s)$ that lies in the 2D or 3D workspace $\mathbb{R}^2$ or $\mathbb{R}^3$. Each edge $e$ has a set of intermediates $e_i$, which are points along the edge. 
The workspace skeleton is a representation of the obstacle-free workspace. 
However, because we often calculate the skeleton with computational geometry methods that create approximations, there are no guarantees that the skeleton lies fully in the obstacle-free workspace. 

\section{Hierarchical Annotated Skeleton RRT}
\label{sec:method}
In Section \ref{sec:hasrrt} we describe the \hasrrtlong\ method and how it works. In Section \ref{sec:adaptingskeletonquality} we discuss the parts of the \hasrrt~algorithm that allow it to adapt to the quality of the workspace skeleton. 

\begin{figure*}[h!]
\centering
\begin{subfigure}{.24\textwidth}
  \centering
  \includegraphics[width=1\textwidth]{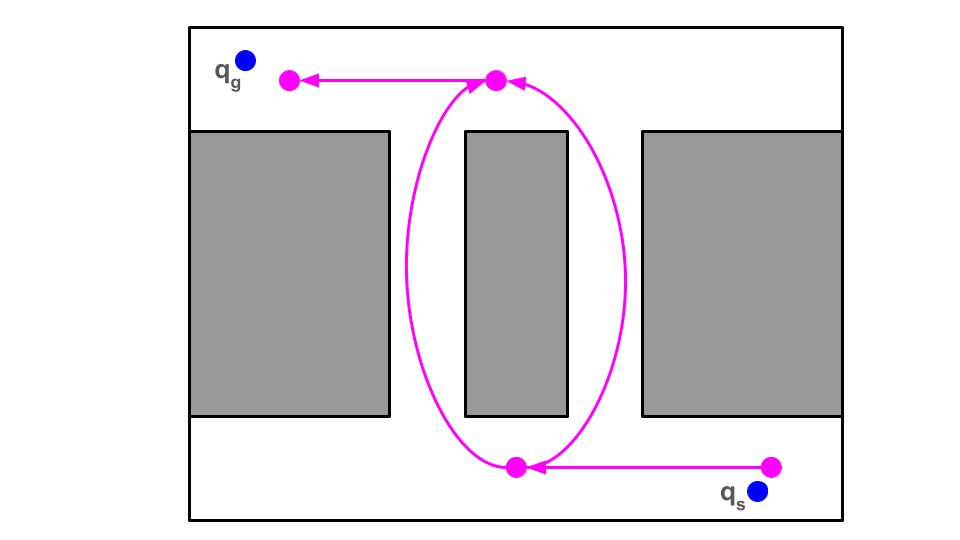}
  \caption{}
  \label{fig:meth1-step1}
\end{subfigure}
\centering
\begin{subfigure}{.24\textwidth}
  \centering
  \includegraphics[width=1\textwidth]{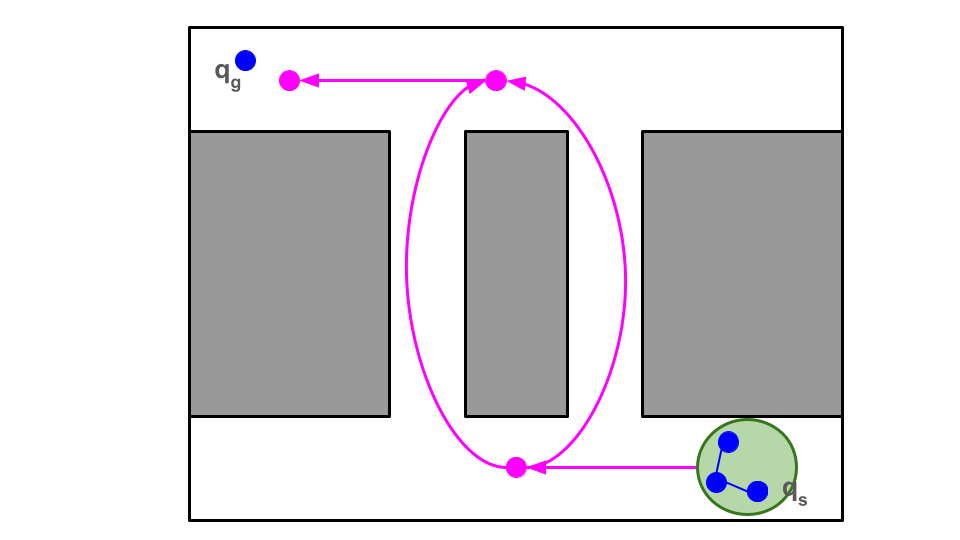}
  \caption{}
  \label{fig:meth1-step2}
\end{subfigure}
\centering
\begin{subfigure}{.24\textwidth}
  \centering
  \includegraphics[width=1\textwidth]{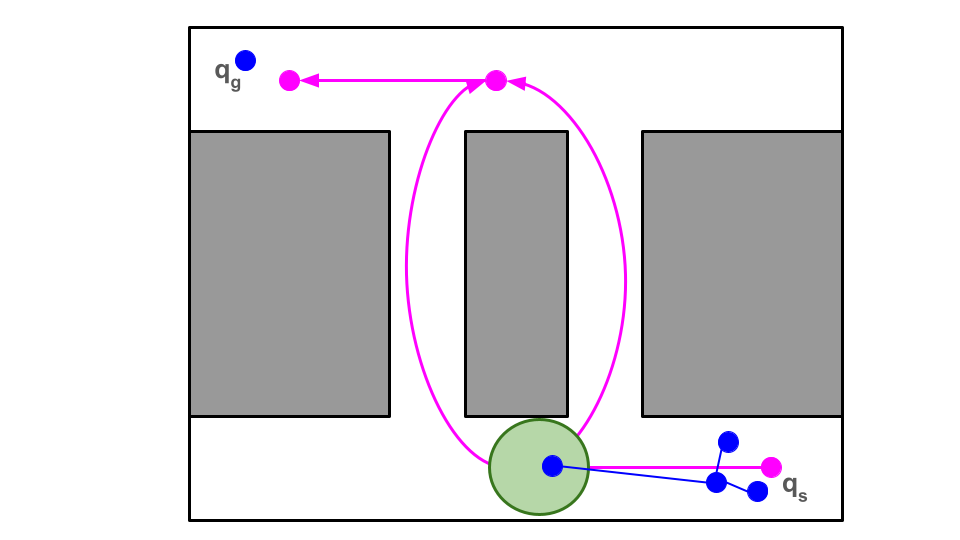}
  \caption{}
  \label{fig:meth1-step3}
\end{subfigure}
\centering
\begin{subfigure}{.24\textwidth}
  \centering
  \includegraphics[width=1\textwidth]{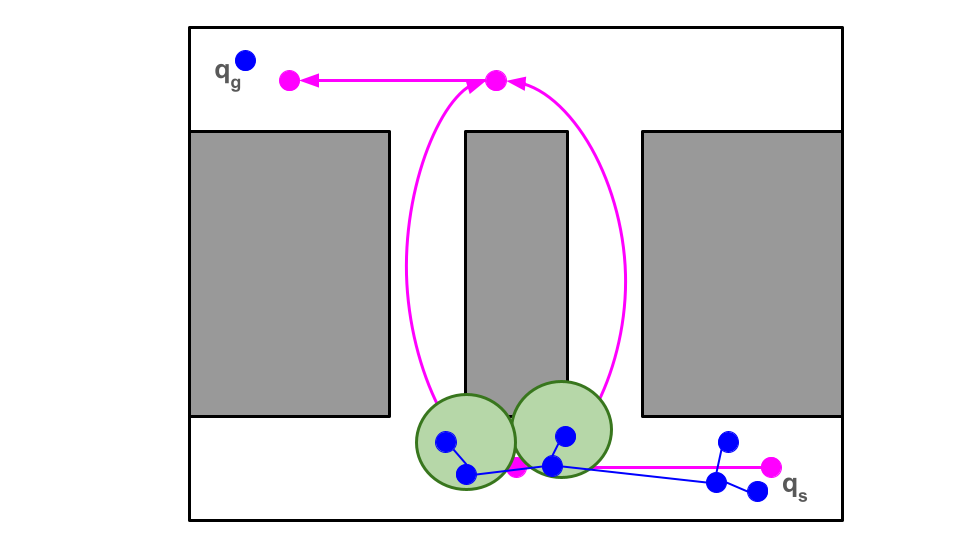}
  \caption{}
  \label{fig:meth1-step4}
\end{subfigure}
\centering
\begin{subfigure}{.24\textwidth}
  \centering
  \includegraphics[width=1\textwidth]{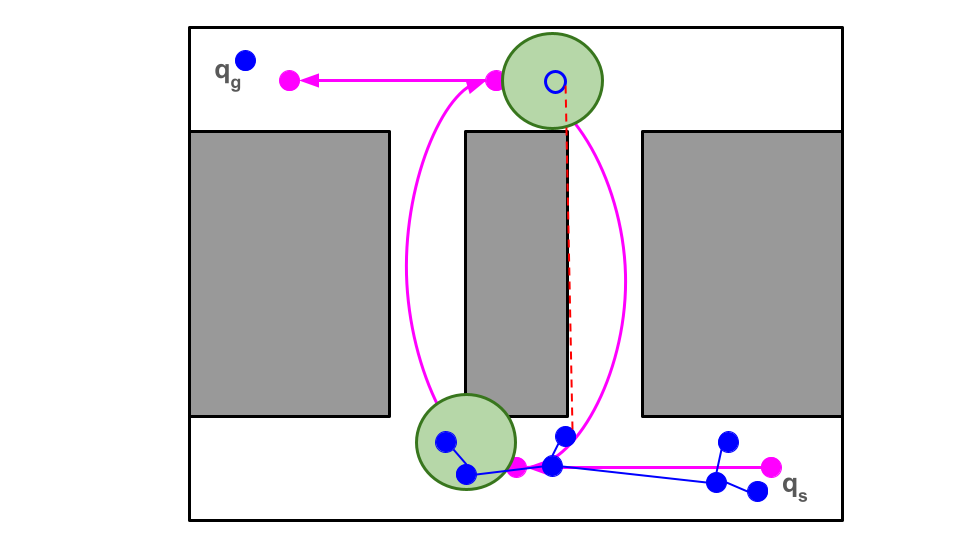}
  \caption{}
  \label{fig:meth1-step5}
\end{subfigure}
\centering
\begin{subfigure}{.24\textwidth}
  \centering
  \includegraphics[width=1\textwidth]{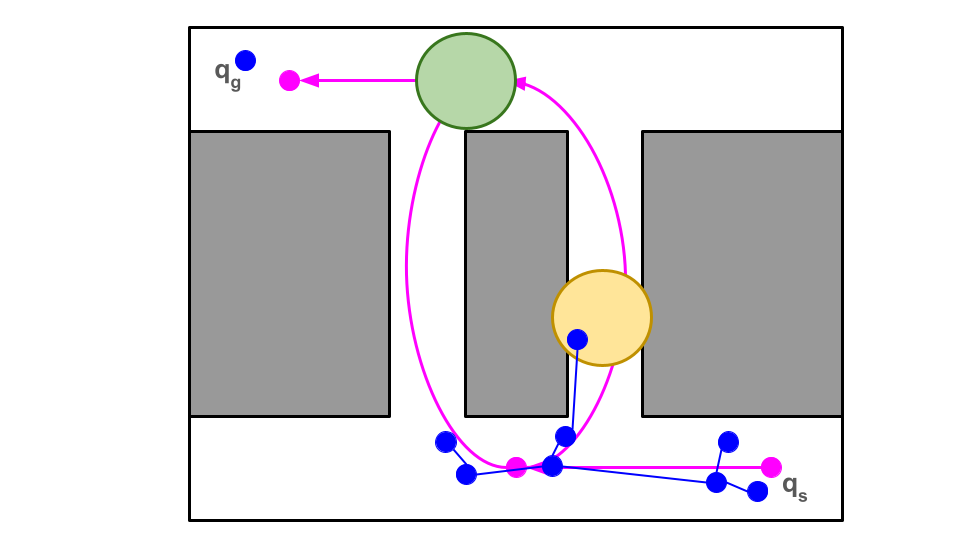}
  \caption{}
  \label{fig:meth1-step6}
\end{subfigure}
\centering
\begin{subfigure}{.24\textwidth}
  \centering
  \includegraphics[width=1\textwidth]{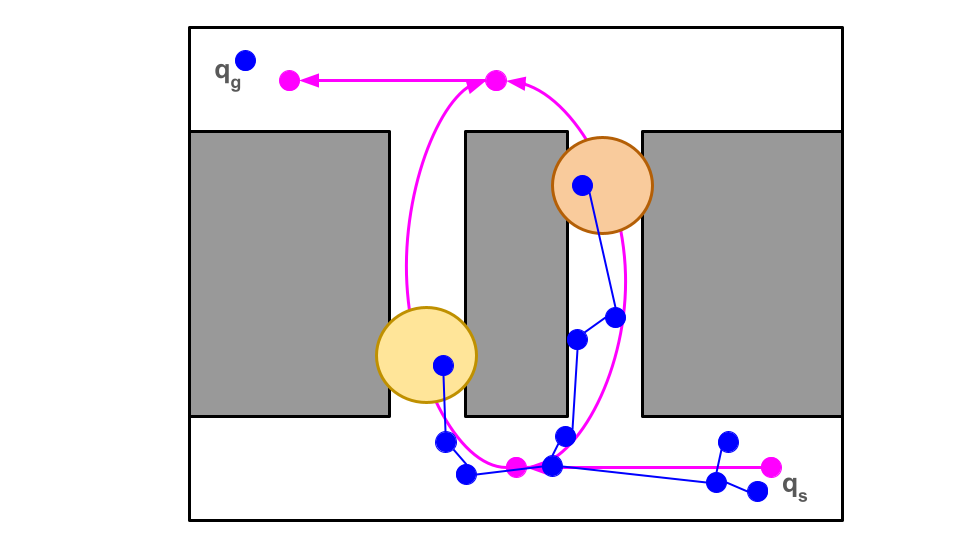}
  \caption{}
  \label{fig:meth1-step7}
\end{subfigure}
\centering
\begin{subfigure}{.24\textwidth}
  \centering
  \includegraphics[width=1\textwidth]{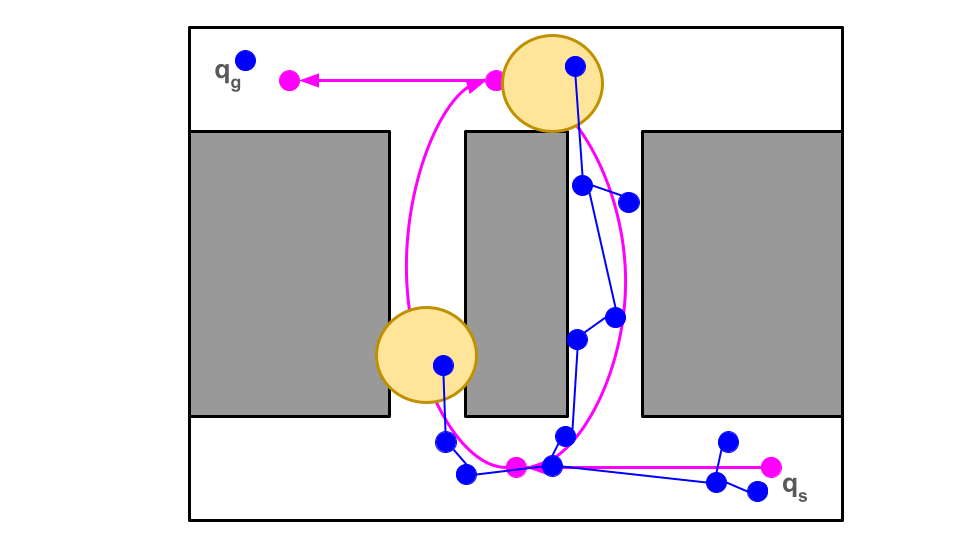}
  \caption{}
  \label{fig:meth1-step8}
\end{subfigure}
\caption{Example execution of the Hierarchical skeleton-guided RRT: (a) Query Skeleton (purple) computed from a query {$q_s$, $q_g$}; (b) initial region (green) placed at the source
vertex of the Query Skeleton; (c) a full extension to the end of the skeleton edge; (d) region splits into two regions at the adjacent edges; (e) an invalid extension is attempted; (f) after a failed extension, a local expansion is attempted; (g) the weight of the failed region is adjusted to bias the tree to alternative path options; (h) the tree after a few iterations.}
 \label{fig:HASRRT}
\end{figure*}

\subsection{The Algorithm}
\label{sec:hasrrt}
The \hasrrtlong~(\hasrrt) algorithm (Figure \ref{fig:HASRRT} and Algorithm \ref{alg:fullalg}) improves upon the previous approach of \drrrt\ by combining exploitation of the skeleton's guidance with local exploration to better adapt to the nature of the surrounding environment. Using the connectivity mapped by the workplace skeleton, \hasrrt\ aims to use longer edge extensions and smaller local extensions when the environment is has more obstacles or the skeleton is less reliable. This allows the algorithm to prioritize paths directly denoted by the skeleton and add exploration only as necessary.
 
The inputs to the algorithm are the 3D workspace, the start configuration $\qstart$, the goal configuration $\qgoal$, and a workspace skeleton $G_s$.


\begin{algorithm}[htb]
 \caption{Hierarchical Annotated Skeleton RRT}
 \label{alg:fullalg}
 \begin{algorithmic}[1]
 \renewcommand{\algorithmicrequire}{\textbf{Input:}}
 \renewcommand{\algorithmicensure}{\textbf{Output:}}
 \REQUIRE Environment $env$, Start $s$, Goal $g$, Annotated Skeleton $aws$
 \STATE{$daws \leftarrow {\tt DirectAndPruneSkeleton}(aws, s)$} \label{alg:directskeleton}
 \STATE{$tree \leftarrow s$} \label{alg:inittree}
 \STATE{$activeRegions \leftarrow {\tt InitializeRegion}(daws, s)$} \label{alg:initregion}
 \WHILE{$!{\tt PathFound}(\qstart, \qgoal)$} 
 \STATE{$r \leftarrow {\tt SelectRegion}(activeRegions, daws)$} \label{alg:selectregion}
 \STATE{$q_{rand} \leftarrow {\tt Sample}(r)$} \label{alg:sample}
 \STATE{$q_{near} = q_{new} = NULL $}
 \STATE{${\tt AttemptExtension}(q_{rand}, tree, q_{near}, q_{new})$}
 \IF{$q_{new} \neq NULL~ OR~ q_{new} \in r$} \label{alg:attemptextension}
 \STATE{$r.{\tt AdvanceRegion}(daws)$} \label{alg:advanceregion}
  \STATE{$r.{\tt IncrementSuccess}()$} \label{alg:incrementsuccess}
\ELSE \label{alg:extensionfailed}
 \STATE{$r.{\tt RetractRegion}(daws, q_{near}, q_{new})$}
 \STATE{$r.{\tt IncrementFailure}()$}
\label{alg:retractregion}
 \ENDIF
 \STATE{$region.{\tt UpdateWeight}()$} \label{alg:updateweight}
 \ENDWHILE
 \end{algorithmic}
 \end{algorithm}

First, the skeleton is directed and pruned to show only the workspace paths relevant to the given query. 
This process is the same as that of \cite{dsba-drbrrt-16}.
The directed skeleton encodes information about the exploration direction should be prioritized as shown in Figure \ref{fig:meth1-step1}.
Since the edges that are deleted during the pruning process do not connect those positions in workspace, they could not connect them in the configuration space either \cite{suda-tgrcdrs-20}. Thus, this step effectively reduces the amount of exploration done by the planner without the possibility of accidentally removing valid solutions. 

\textit{Initializing the tree} (Lines \ref{alg:inittree} and \ref{alg:initregion} in Algorithm \ref{alg:fullalg}):
We define a sampling region as a sphere in the $\cspace$ of radius $r$ (determined through user-selected hyperparameters), anchored to an intermediate $e_i$. 
An active sampling region $s$ is initialized at the skeleton vertex nearest to $\qstart$ in the workspace. The tree is initialized with the start configuration $\qstart$ and grown until it reaches the first sampling region. Figure \ref{fig:meth1-step2} illustrates the initial region with a tree of size 3.

\textit{Expanding the tree} (Lines \ref{alg:selectregion} to \ref{alg:updateweight} in Algorithm \ref{alg:fullalg}): These steps constitute the main loop that terminates when the query is solved or the algorithm runs out of resources. 
First, an active region is selected (see Section \ref{sec:adaptingskeletonquality} for region selection details), and the planner samples inside for a direction to expand the tree toward. 
Once a valid $\qrand$ is sampled, an extension attempt is performed to expand the tree (Line \ref{alg:attemptextension} of Algorithm \ref{alg:fullalg}). 
The successful extension adds a new vertex to the tree and pushes the region that gave $\qrand$ to the end of its current skeleton edge. 
The region's weight is also updated by incrementing its success record to increase its chances of being selected in the next iterations. 
Then, new regions are created at the start of all outgoing new skeleton edges.
A successful direct extension to the end of a skeleton edge and its branching into two regions for each adjacent edge are shown in Figures \ref{fig:meth1-step3} and \ref{fig:meth1-step4}.

\textit{Retracting a sampling region} (Lines \ref{alg:extensionfailed}-\ref{alg:retractregion} in Algorithm \ref{alg:fullalg}):
If the extension does not reach the region where $q_{rand}$ was sampled, as shown in Figure \ref{fig:meth1-step5}, the region is pulled back toward $\qnear$. The current implementation pulls the region halfway between $\qrand$ and $\qnear$ as shown in Algorithm \ref{alg:retract-region} and Figure \ref{fig:meth1-step6}. In addition, the region's weight is updated by incrementing its failure record to decrease its chances of being selected in the next iterations, as shown in Figure \ref{fig:meth1-step7}.

\begin{algorithm}[t!]
\caption{Region Retraction}
\label{alg:retract-region}
\begin{algorithmic}[2]
    \REQUIRE Region $r$, Current annotated skeleton $ask$, $q_{near}$, $q_{new}$
    \STATE{$edge \leftarrow ask.{\tt GetSkeletonEdge}(r)$}
    \STATE{$currPos \leftarrow r.{\tt GetCenter}()$}
    \IF{$q_{new} \neq NULL$}
    \STATE{$prevPos \leftarrow q_{new}$}
    \ELSE
    \STATE{$prevPos \leftarrow q_{near}$}
    \ENDIF
    \STATE{$newPos \leftarrow {\tt GetMidPoint}(prevPos, currPos)$}
    \STATE{$r.center \leftarrow newPos$}
    \RETURN{$-({\tt Distance}(newPos, currPos))$}
\end{algorithmic}
\end{algorithm}

\textit{Region Selection.} 
The planner chooses the next region using a probability distribution based on prior extension success, the directed skeleton, and the explore/exploit bias. 

A candidate region $r$ is selected with probability 
$p_r$ calculated as follows, where $e$ is the explore bias hyperparameter and $R$ is the set of all regions:
$$p_r = \frac{e}{|R|+1} + (1-e)\frac{w_r}{\sum_{r'\in R}w_{r'}}$$

A region's weight $w_r$ is the percentage of successful extensions within that region. 
To maintain probabilistic completeness, the whole environment also constitutes one region in case the workspace skeleton does not cover parts of the environment that comprise the motion planning solution. 
The environment is chosen with probability $\frac{e}{|R|+1}$.
When the whole environment is selected a sample is chosen randomly within the environment, and the planner behaves like a regular RRT.

\subsection{Adapting to Skeleton Quality - Method}
\label{sec:adaptingskeletonquality}
Strategies guided by the workspace skeleton are affected by the quality of the skeleton. In \drrrt\, when a skeleton edge does not provide reliable guidance, the sampling region on the edge leads to repeated sampling failure, which lowers its weight and eventually pushes the algorithm to follow other directions if available or degrade to random sampling of the whole environment. 
With  \hasrrt\, the effects of poor guidance are mitigated by three things. First, the skeleton annotations help prioritize edges that lead to the desired solution, making it less likely to encounter bad guidance. 
Second, when a direct extension along the skeleton edge fails, the algorithm's retraction mechanism keeps the exploration of an alternative focused on adjusting the extension at the point of failure instead of directly reverting to random sampling. 
Third, if the guidance provided by the skeleton is so poor that the algorithm experiences increasingly more failures than successes by using the skeleton, the target selection behavior reverts back to that of \rrt, allowing for randomized exploration of the environment.

Prior extension success denotes extending to a sample within a region $r$. If the skeleton is reliable, then $r$ on a skeleton vertex $v$ is centered properly in $\cfree$. Thus, the algorithm is more likely to have repeated success extending into $r$ again. The directed skeleton ensures that the algorithm continues progressing toward its goal. If a particular skeleton vertex is not accurately placed in the free workspace, the samples in the corresponding $\cspace$ may not be valid as often. Thus, the success rate in $r$ goes down, making the region less desirable to expand into. 
Future region selections will prioritize other regions with higher success rates over $r$.

\section{Validation}
\label{sec:validation} 

\begin{figure}[t]
\begin{subfigure}{.23\textwidth}
  \centering
  \includegraphics[width=1\textwidth]{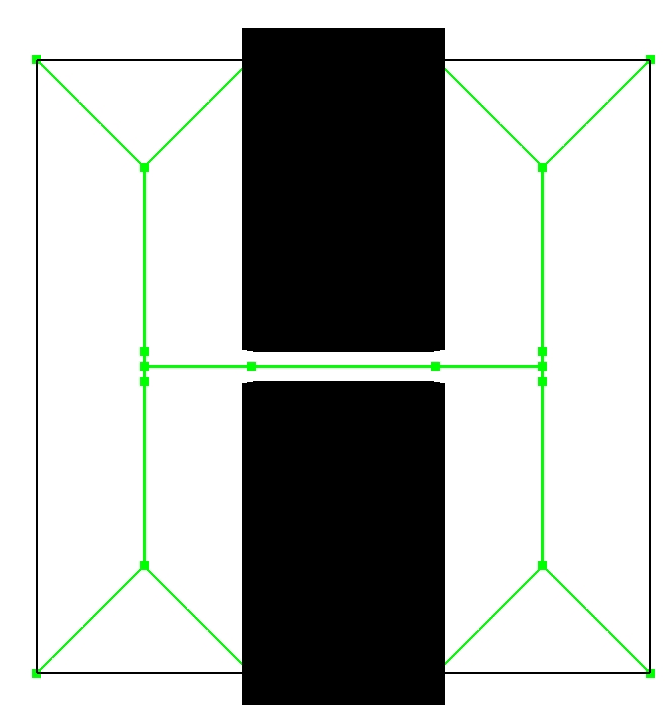}
  \caption{2D narrow passage and Medial Axis skeleton. We use a small 3-DOF rectangular prism robot. 
  The robot at its widest is 60\% of narrow passage's width.
  }
  \label{fig:envs-simplepassage}
\end{subfigure}
\begin{subfigure}{.23\textwidth}
  \centering
  \includegraphics[width=1\textwidth]{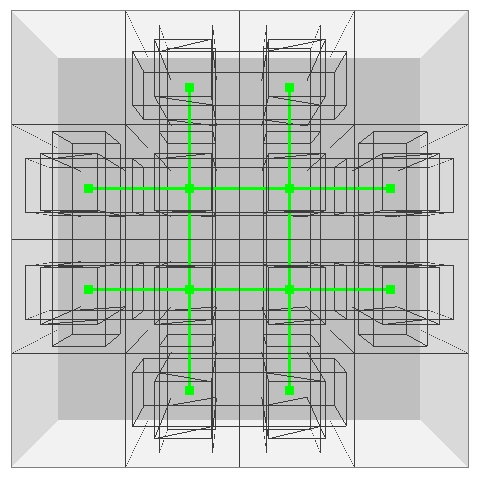}
  \caption{Grid tunnels and skeleton constructed with method from Section \ref{sec:skeleton-construction}. We use a small 6-DOF L-shaped robot. The L shape is approximately 8\% of a block's width.} 
  \label{fig:envs-gridtunnels}
\end{subfigure}
\begin{subfigure}{.46\textwidth}
  \centering
  \includegraphics[width=1\textwidth]{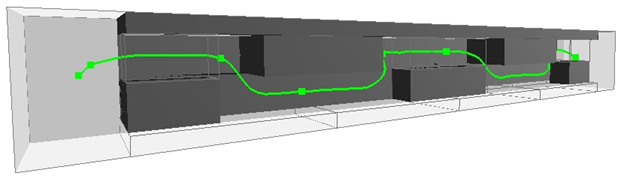}
  \caption{Extended Z-shaped tunnels and a curated mean curvature skeleton. We use a cube-shaped robot that cannot rotate (3-DOF). The robot's width is approximately 10\% of the tunnel's width.}
  \label{fig:envs-extendedz}
\end{subfigure}
\begin{subfigure}{.23\textwidth}
  \centering
  \includegraphics[width=1\textwidth]{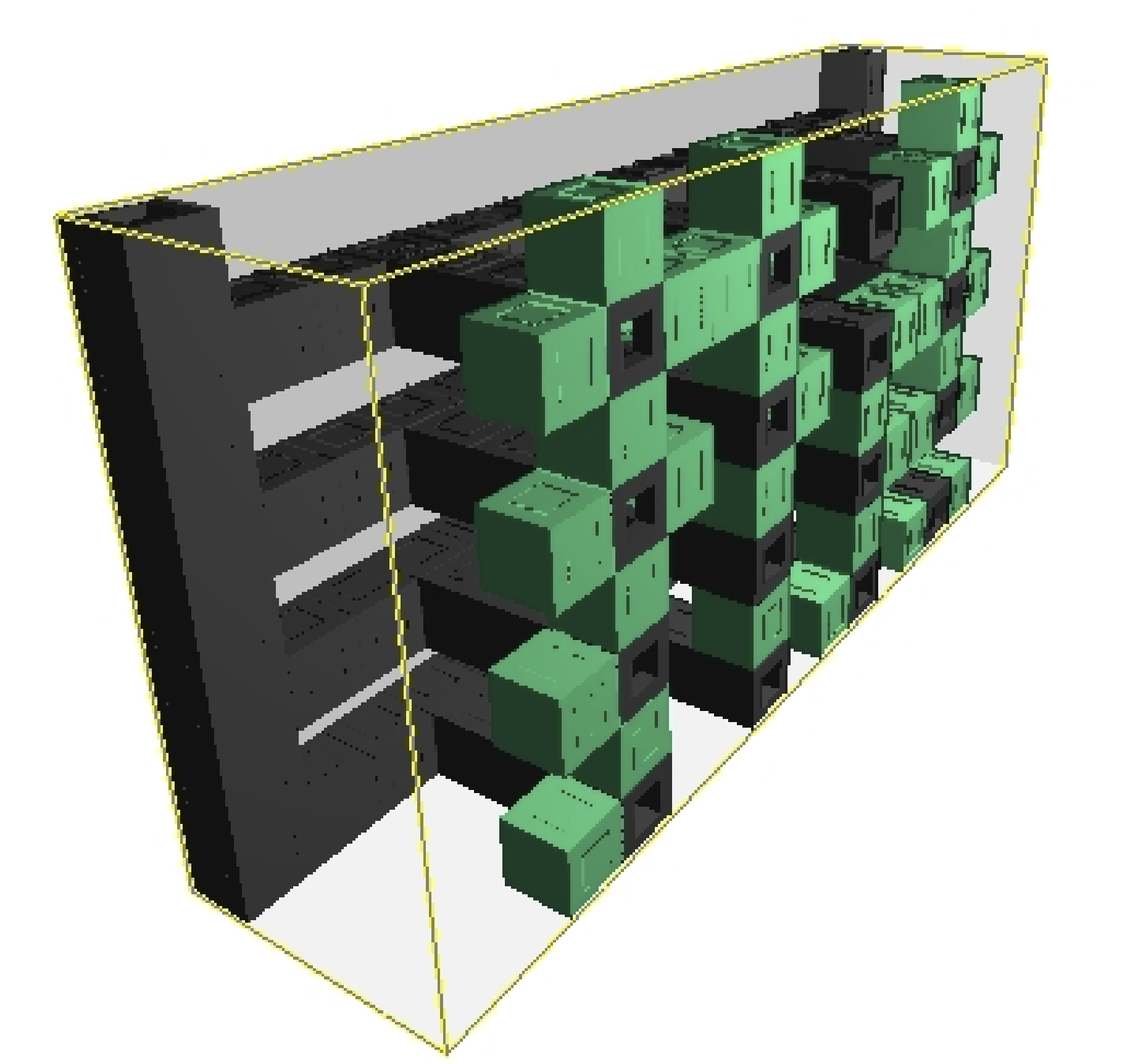}
  \caption{Grid Mining, with shafts (grey) and drifts (green), and skeleton constructed with method from Section \ref{sec:skeleton-construction}. We use a small 6-DOF L-shaped robot. The L shape is approximately 8\% of one block's width.} 
  \label{fig:envs-gridmine}
\end{subfigure}
\begin{subfigure}{.23\textwidth}
  \centering
  \includegraphics[width=1\textwidth]{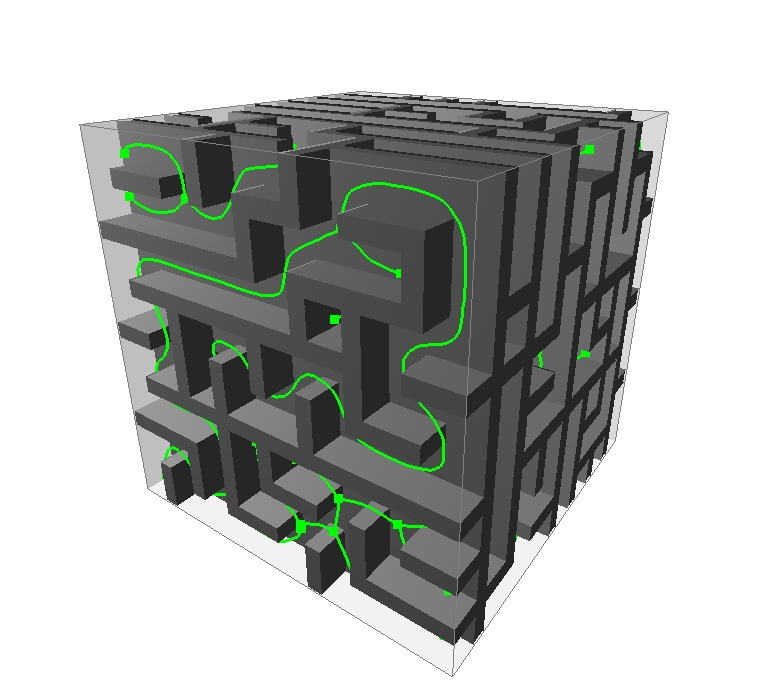}
  \caption{8x8 3D grid maze and curated mean curvature skeleton. We use a stick shaped robot that can rotate around any axis 6-DOF. The length and depth of the stick are approximately 2\% and 9\% of the environment's length, respectively.}
  \label{fig:envs-gridmaze}
\end{subfigure}
\caption{Tested environments. Workspace skeletons are shown in green. }
\label{fig:envs}
\end{figure}

\label{sec:setup}
We compare \hasrrt~with three other methods: \basicrrt \cite{lk-rrtpp-00}, \eet \cite{Rickert_balancingexploration}, and \drrrt \cite{dsba-drbrrt-16}. \drrrt\ and \eet\ use workspace guidance to help with sampling. These methods were chosen due to their ability to grow an RRT with additional environmental guidance. We compare with RRT because it is a baseline method. 

The results below show the robustness of \hasrrt\ to all environments. 
In each environment, one of the other planning strategies' performance is comparable to that of \hasrrt\, but only \hasrrt\ consistently yields good solutions in less time than the others. In addition, \hasrrt\ is independent of parameter tuning, an added value that makes it a good choice for a non-expert user.

We evaluate the method in five maze/tunnel environments shown and described in Figure \ref{fig:envs}.
In Section \ref{sec:experimental-setup} and  \ref{sec:skeleton-construction} we describe our experimental setup and our novel modular way to construct a workspace skeleton for two of our environments. In Section \ref{sec:discussion} we discuss the results of our experiments. Finally, in Section \ref{sec:ablation-study} we analyze the performance of \hasrrt~with varying qualities of guidance to demonstrate how \hasrrt~behaves with poor guidance.

\subsection{Experimental Setup}
\label{sec:experimental-setup}
In each environment, a query was set to evaluate the ability of the \rrt~ variant to solve an accessibility problem in that environment.
We pre-computed different types of skeletons and annotated them
with clearance for all the problems.
The results do not report the time to compute a workspace skeleton since they are incurred
only once. However, the time to prune and direct the skeleton is reported in the run time. Each algorithm was given the same amount of time to solve the
queries, ranging from 20 seconds to 300, depending on the problem's difficulty. 
We report the time to solve the query and path cost. For robotics environments, we use path
length as a proxy for path cost.
For all planners and in all environments, we use Euclidean distance as a distance metric, and our local planner always moves in a straight line towards its goal. 

The experiments were executed on a Google Cloud Compute e2-standard-4 computer with 4vCPUs and 16GB of RAM, running Ubuntu 20.04.
All methods were implemented in our C++ Parasol Planning
Library (PPL) \cite{PPL}. 
Validation was done with
PQP-SOLID collision detection \cite{lglm-pqp-99}. 
Hyperparameters 
were chosen through experimentation.

\subsection{Skeleton Construction and Annotation}
\label{sec:skeleton-construction}

\begin{figure}[h!]
\begin{subfigure}{0.5\textwidth}
    \centering
    \includegraphics[width=0.65\textwidth]{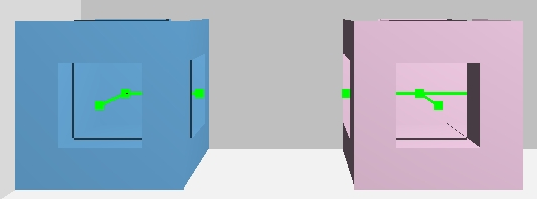}
    \captionsetup{width=0.8\textwidth}
    \caption{Two separate blocks, blue with two entryways and pink with three entryways. Two separate skeletons. }
    \label{fig:blocks_separate}    
\end{subfigure}
\begin{subfigure}{0.5\textwidth}
    \centering
    \includegraphics[width=0.4\textwidth]{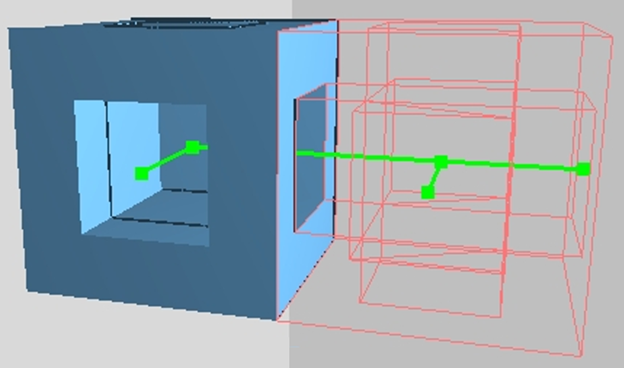}
    \captionsetup{width=0.8\textwidth}
    \caption{Blue (solid) and pink (outlined) blocks, combined, with pruned skeleton. }
    \label{fig:blocks_combined}
\end{subfigure}
\caption{Block environment and skeleton construction.}
\label{fig:blocks}
\end{figure}

The workspace skeleton is constructed differently depending on the environment. For example, a medial-axis skeleton can easily be constructed for a 2-dimensional environment like the 2D narrow passage in Figure \ref{fig:envs-simplepassage}. In 3D cases like the extended Z tunnel and the grid maze, structures like the mean curvature skeleton \cite{blk-tcsrp-2012} can be used to show the connectivity of the workspace. In such environments, however, the default graph may contain long curved edges that do not indicate the changes in the topology of the workspace. For example, a tunnel with a $90^{\circ}$ turn may be represented by a long curved edge with no indication of the change in direction.

In some cases, instead of constructing an environment and computing a skeleton afterward, we simultaneously build an environment and skeleton. For simplicity, we create an environment with $90^{\circ}$ angles using cubic structures as building blocks.
Considering orientation, there are $2^6$ ways to construct a cube with varying tunnels leading outwards.
After designing the blocks, a skeleton is constructed for each block based on the tunnel openings, as shown in Figure \ref{fig:blocks_separate}. When blocks are paired, their respective skeletons are joined to make a connected graph representing the new topology (Figure \ref{fig:blocks_combined}).

\subsection{Discussion}
\label{sec:discussion}

\begin{table*}[htb]
\centering
\caption{Graph size, success rate, and planning cost comparisons for the tested environments. For seeds which do not find paths within the specified time, we include the number of vertices and collision detection calls for the tree built in the allotted time. We do not run \rrt~and \eet~in the Grid Maze environment due to the environment's size and complexity. }
\label{tab:results}
\begin{adjustbox}{width=\textwidth}
\begin{tabular}{|l|l|rrrr|rrrr|rrrr|}
\hline
\multirow{2}{*}{Environment}       &     & \multicolumn{4}{c|}{Number of vertices}                                                                  & \multicolumn{4}{c|}{Collision Detection Calls}                                                                 & \multicolumn{4}{c|}{Completed seeds ($\%$)}                                                                                                                     \\ \cline{3-14} 
                                   &     & \multicolumn{1}{r|}{RRT} & \multicolumn{1}{r|}{EET}         & \multicolumn{1}{r|}{DR-RRT} & HAS-RRT      & \multicolumn{1}{r|}{RRT}   & \multicolumn{1}{r|}{EET}           & \multicolumn{1}{r|}{DR-RRT} & HAS-RRT        & \multicolumn{1}{r|}{RRT}                    & \multicolumn{1}{r|}{EET}                   & \multicolumn{1}{r|}{DR-RRT}                 & HAS-RRT                \\ \hline
\multirow{2}{*}{Simple Passage}    & avg & \multicolumn{1}{r|}{462} & \multicolumn{1}{r|}{660}         & \multicolumn{1}{r|}{59}     & \textbf{20}  & \multicolumn{1}{r|}{6138}  & \multicolumn{1}{r|}{6020}          & \multicolumn{1}{r|}{725}    & \textbf{393}   & \multicolumn{1}{r|}{\multirow{2}{*}{100\%}} & \multicolumn{1}{r|}{\multirow{2}{*}{63\%}} & \multicolumn{1}{r|}{\multirow{2}{*}{100\%}} & \multirow{2}{*}{100\%} \\ \cline{2-10}
                                   & std & \multicolumn{1}{r|}{425} & \multicolumn{1}{r|}{833}         & \multicolumn{1}{r|}{14}     & 4            & \multicolumn{1}{r|}{4418}  & \multicolumn{1}{r|}{6506}          & \multicolumn{1}{r|}{142}    & 56             & \multicolumn{1}{r|}{}                       & \multicolumn{1}{r|}{}                      & \multicolumn{1}{r|}{}                       &                        \\ \hline
\multirow{2}{*}{Grid Tunnels}      & avg & \multicolumn{1}{r|}{86}  & \multicolumn{1}{r|}{\textbf{16}} & \multicolumn{1}{r|}{438}    & 19           & \multicolumn{1}{r|}{5383}  & \multicolumn{1}{r|}{\textbf{1359}} & \multicolumn{1}{r|}{8000}   & 1496           & \multicolumn{1}{r|}{\multirow{2}{*}{100\%}} & \multicolumn{1}{r|}{\multirow{2}{*}{97\%}} & \multicolumn{1}{r|}{\multirow{2}{*}{100\%}} & \multirow{2}{*}{100\%} \\ \cline{2-10}
                                   & std & \multicolumn{1}{r|}{61}  & \multicolumn{1}{r|}{3}           & \multicolumn{1}{r|}{78}     & 2            & \multicolumn{1}{r|}{2866}  & \multicolumn{1}{r|}{353}           & \multicolumn{1}{r|}{1103}   & 305            & \multicolumn{1}{r|}{}                       & \multicolumn{1}{r|}{}                      & \multicolumn{1}{r|}{}                       &                        \\ \hline
\multirow{2}{*}{Extended Z Tunnel} & avg & \multicolumn{1}{r|}{592} & \multicolumn{1}{r|}{\textbf{51}} & \multicolumn{1}{r|}{498}    & 55           & \multicolumn{1}{r|}{15922} & \multicolumn{1}{r|}{3922}          & \multicolumn{1}{r|}{7288}   & \textbf{1977}  & \multicolumn{1}{r|}{\multirow{2}{*}{100\%}} & \multicolumn{1}{r|}{\multirow{2}{*}{94\%}} & \multicolumn{1}{r|}{\multirow{2}{*}{100\%}} & \multirow{2}{*}{100\%} \\ \cline{2-10}
                                   & std & \multicolumn{1}{r|}{199} & \multicolumn{1}{r|}{13}          & \multicolumn{1}{r|}{60}     & 7            & \multicolumn{1}{r|}{4163}  & \multicolumn{1}{r|}{559}           & \multicolumn{1}{r|}{686}    & 136            & \multicolumn{1}{r|}{}                       & \multicolumn{1}{r|}{}                      & \multicolumn{1}{r|}{}                       &                        \\ \hline
\multirow{2}{*}{Grid Mining}       & avg & \multicolumn{1}{r|}{634} & \multicolumn{1}{r|}{\textbf{78}} & \multicolumn{1}{r|}{1701}   & 130          & \multicolumn{1}{r|}{52102} & \multicolumn{1}{r|}{8682}          & \multicolumn{1}{r|}{29114}  & \textbf{5750}  & \multicolumn{1}{r|}{\multirow{2}{*}{31\%}}  & \multicolumn{1}{r|}{\multirow{2}{*}{80\%}} & \multicolumn{1}{r|}{\multirow{2}{*}{100\%}} & \multirow{2}{*}{100\%} \\ \cline{2-10}
                                   & std & \multicolumn{1}{r|}{136} & \multicolumn{1}{r|}{40}          & \multicolumn{1}{r|}{113}    & 2            & \multicolumn{1}{r|}{9345}  & \multicolumn{1}{r|}{1519}          & \multicolumn{1}{r|}{1488}   & 155            & \multicolumn{1}{r|}{}                       & \multicolumn{1}{r|}{}                      & \multicolumn{1}{r|}{}                       &                        \\ \hline
\multirow{2}{*}{Grid Maze}         & avg & \multicolumn{1}{r|}{NaN} & \multicolumn{1}{r|}{NaN}         & \multicolumn{1}{r|}{3570}   & \textbf{892} & \multicolumn{1}{r|}{NaN}   & \multicolumn{1}{r|}{NaN}           & \multicolumn{1}{r|}{69675}  & \textbf{26124} & \multicolumn{1}{r|}{\multirow{2}{*}{NaN}}   & \multicolumn{1}{r|}{\multirow{2}{*}{NaN}}  & \multicolumn{1}{r|}{\multirow{2}{*}{97\%}}  & \multirow{2}{*}{100\%} \\ \cline{2-10}
                                   & std & \multicolumn{1}{r|}{NaN} & \multicolumn{1}{r|}{NaN}         & \multicolumn{1}{r|}{183}    & 197          & \multicolumn{1}{r|}{NaN}   & \multicolumn{1}{r|}{NaN}           & \multicolumn{1}{r|}{3565}   & 4979           & \multicolumn{1}{r|}{}                       & \multicolumn{1}{r|}{}                      & \multicolumn{1}{r|}{}                       &                        \\ \hline
\end{tabular}
\end{adjustbox}
\end{table*}

\begin{figure*}[h!]
\begin{subfigure}{0.5\textwidth}
    \centering
    \includegraphics[width=1\textwidth]{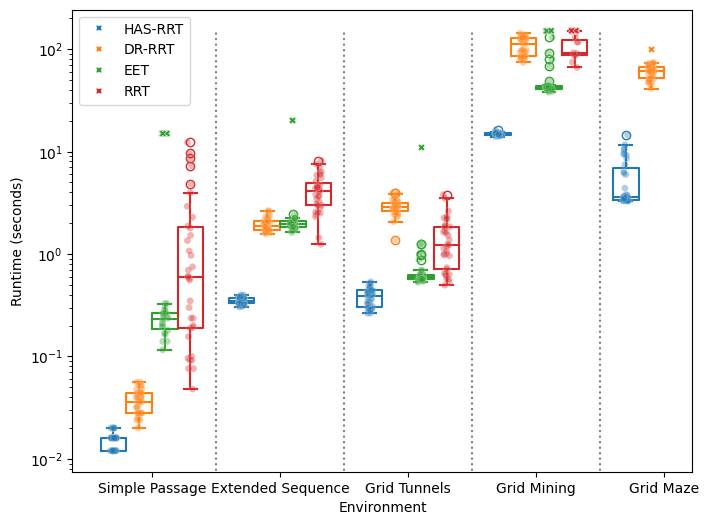}
    \caption{Run time in seconds, for all seeds. }
    \label{fig:results-runtime}    
\end{subfigure}
\begin{subfigure}{0.5\textwidth}
    \centering
    \includegraphics[width=1\textwidth]{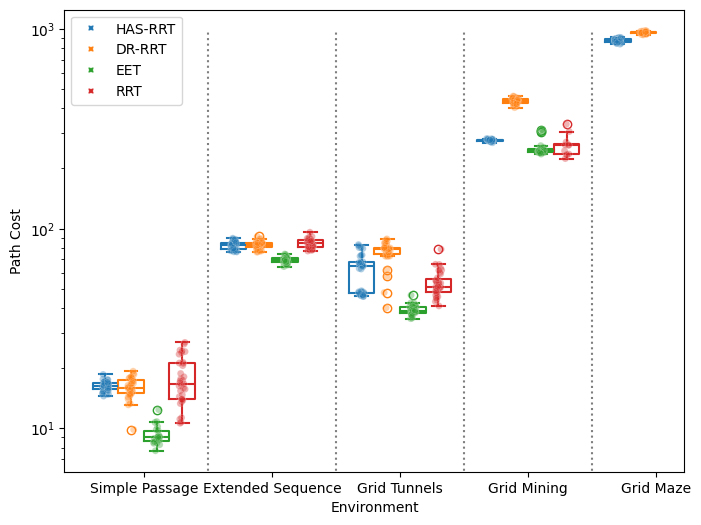}
    \caption{Path cost (path edge weights) for only seeds that found a path.}
    \label{fig:results-pathcost}
\end{subfigure}
\caption{Run time and path cost comparisons for the tested robotics environments.`x' marks represent seeds that failed to find a solution. }
\label{fig:results}
\end{figure*}

The experimental results in Figure \ref{fig:results} and Table \ref{tab:results} show the importance of using a workspace skeleton to guide planning and the significant advantage of using hierarchical planning with guidance. The comparative results show that \drrrt\ takes longer than \hasrrt\ and is not consistently faster than basic \rrt\ or \eet. In addition, although \eet\ returns paths with the lowest cost, its success rate is worse than others in most environments, and it is sensitive to parameter tuning. 

In the Simple Passage, Extended Z Tunnel, and Grid Mining environments, \hasrrt~outperformed the other guided strategies
because of the long extensions it takes through the narrow passage using skeleton guidance. 
\drrrt~locally explores each edge, increasing its sampling failures in the narrow passage. 
\eet~takes longer for the wavefront expansion  to discover and explore the narrow passage randomly, but the resulting guidance yields a better path cost.
These examples highlight one of the main strengths of \hasrrt: prioritizing paths indicated by the workspace guidance and only sampling if a naive extension fails. By following the guidance of a good skeleton, the runtime of \hasrrt\ is greatly minimized while maintaining a competitive path cost.

In the grid tunnel environment, \hasrrt~runs faster than \drrrt~and \eet. \eet~performs faster than \drrrt~because the wavefront expansion is easier to build in this environment. 

The grid mine environment shows the scalability of \hasrrt. In this more complex environment, \basicrrt~only solved 31\% of the 35 random seeds. \hasrrt~even performs better than \drrrt, because it simply follows the guidance from the long skeleton edges rather than fully exploring each edge. 

The grid maze environment is notoriously difficult and was only successfully solved by \drrrt~and \hasrrt~(of the 35 seeds). 
This environment has curved edges (as shown in Figure \ref{fig:envs-gridmaze}), even though the environment's tunnels are straight. 
From this, we can see two more strengths of \hasrrt: 
While \hasrrt~works best with straight-edged workspace skeletons, it performs well on other types of skeletons due to the method's exploration when necessary. 
Additionally, as evidenced by \hasrrt's lower path cost and faster runtime compared to \drrrt, \hasrrt~still uses more information from the skeleton than its competitors. 

\eet's path costs are comparable and competitive to \hasrrt\ because the Wavefront expansion guarantees any path to be centered in an environment.
However, \eet's runtime is much higher than the other methods because the wavefront expansion is built probabilistically and thus must be recomputed for each seed. 
Additional investigation shows that building the wavefront expansion takes up most of the runtime: 
build times for the Generated Grid environment averaged 66.49\% of the total runtime and for the Extended Z Tunnel averaged 85.75\%. The other two environments averaged similarly. 

\subsection{Adapting to Skeleton Quality - Study} \label{sec:ablation-study}

We also investigate \hasrrt's capability to consistently find optimal paths with an unreliable skeleton. 
We ran \hasrrt\ on the Grid Tunnels environment with increasingly unreliable skeletons and compare runtime and path cost. 
Since the skeleton generated using the process from Section \ref{sec:skeleton-construction} is medially centered in the Grid Tunnels, it can be easily modified and its reliability can be easily quantified. 

\begin{figure}[t]
            \begin{subfigure}{0.24\textwidth}
                \centering
                \includegraphics[width=\textwidth, trim={3.5cm, 3cm, 3cm, 3.5cm}, clip]{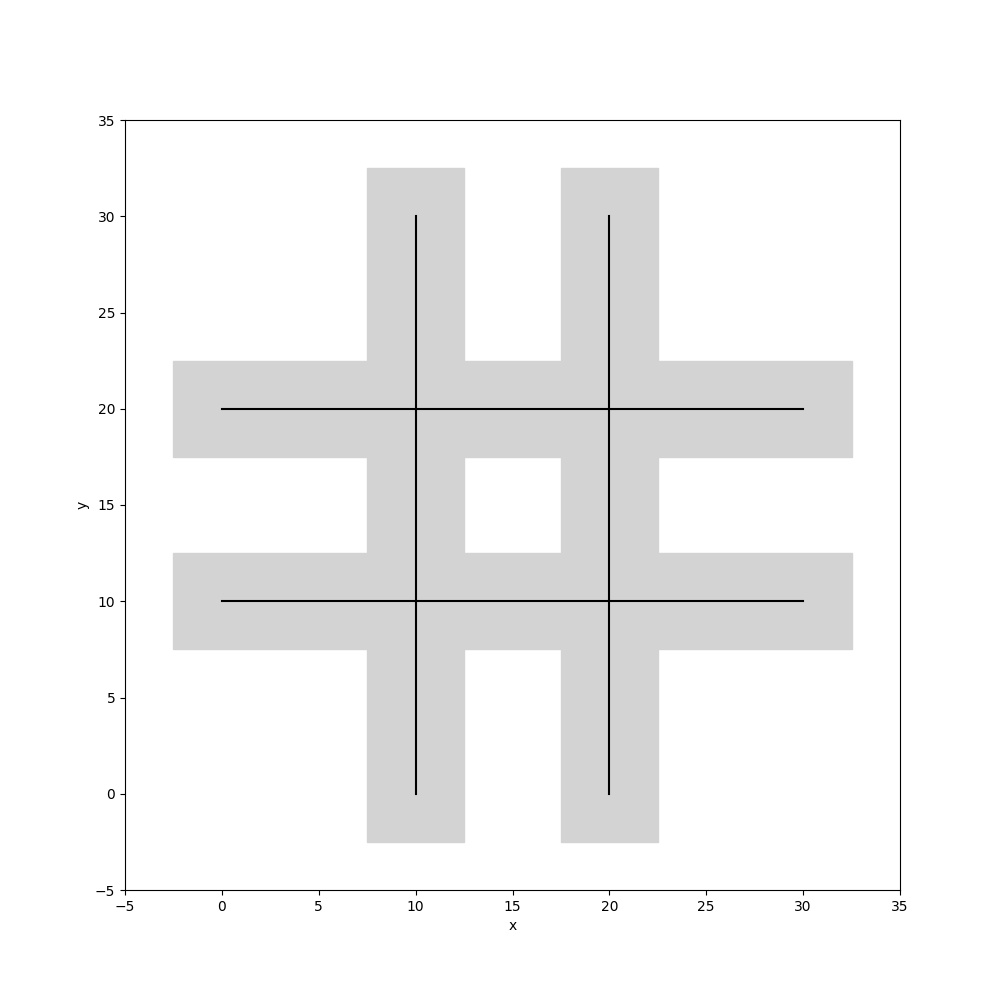}
                \caption{$d=0$}
                \label{fig:skeleton_0}
            \end{subfigure}
            \begin{subfigure}{0.24\textwidth}
                \centering
                \includegraphics[width=\textwidth, trim={3.5cm, 3cm, 3cm, 3.5cm}, clip]{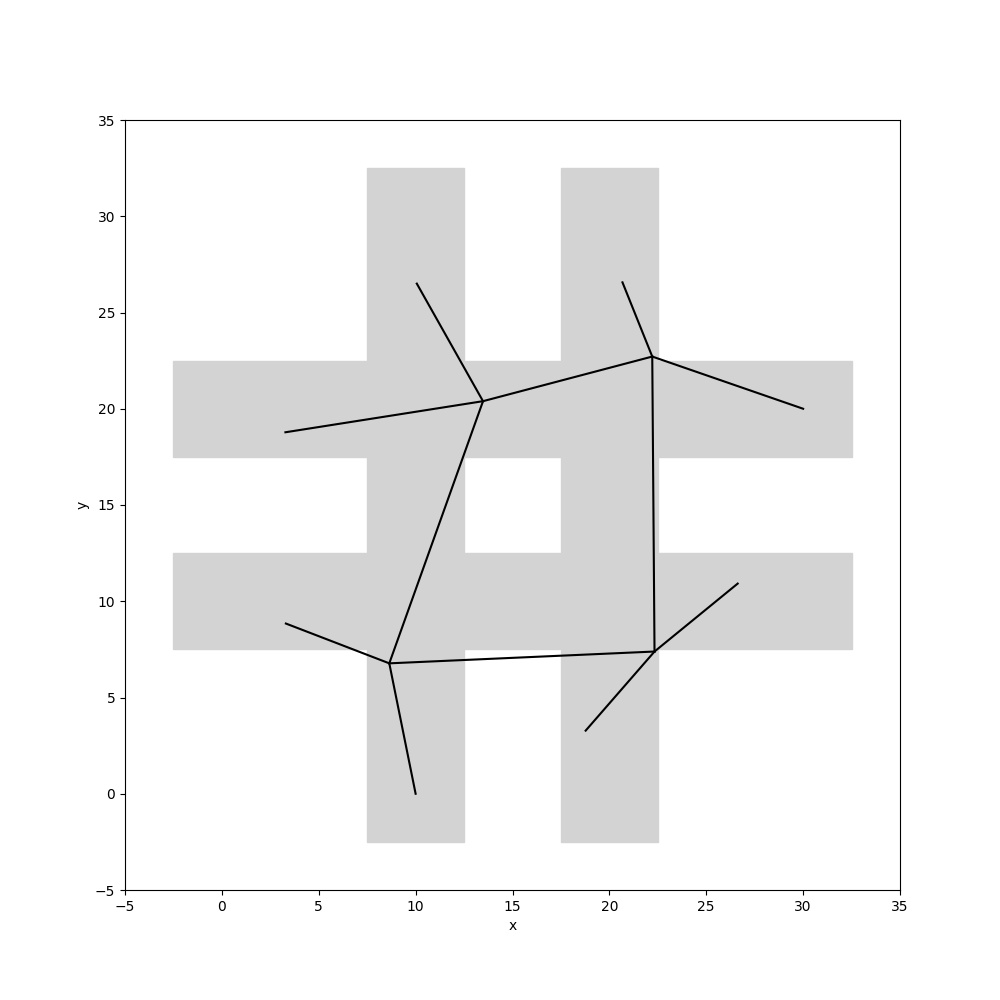}
                \caption{$d=3.5$}
                \label{fig:skeleton_35}
            \end{subfigure}
\caption{Perturbed skeletons. Valid regions for the grid environment are shown in grey, and the skeleton is shown in black. }
\label{fig:perturbed_skeleton_comparison}
\end{figure}

We used the same query used in the experiments from Section \ref{sec:validation}. Each vertex of the skeleton not corresponding to the query start and goal positions are shifted by a distance $d$ in a random direction and are guaranteed to be within the free workspace (see Figure \ref{fig:perturbed_skeleton_comparison}). 
We increase the maximum radius of a sampling region from the previous set of experiments to accommodate the poorer-quality skeletons. 
Each experiment is run with 10 seeds. 

\begin{figure}[t]
\centering
\begin{subfigure}{0.5\textwidth}
    \centering
    \includegraphics[width=1\textwidth]{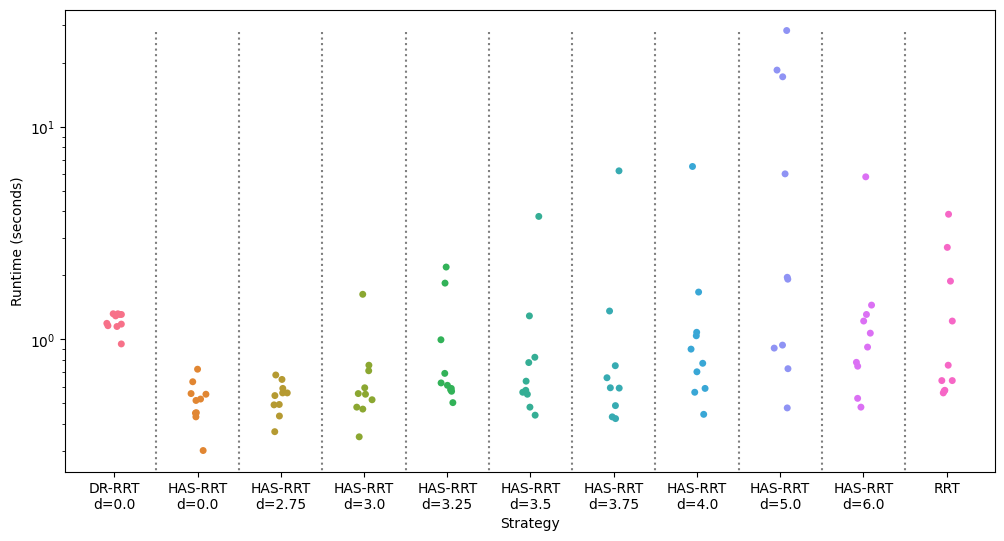}
\end{subfigure}    
\caption{Runtimes for \drrrt, \hasrrt\ for ten different perturbed skeletons, and \rrt. Only runtimes for seeds which found paths are shown. }
\label{fig:perturbed_results}    
\end{figure}

The results for these experiments are shown in Figure \ref{fig:perturbed_results}. The \hasrrt\ results are shown in increasing order of $d$ from the second to tenth columns. Smaller values of $d$ have results that match those in Section \ref{sec:discussion}, where \hasrrt's performance is faster than both \drrrt\ and \rrt. As $d$ increases, the quality of the skeleton worsens, and the algorithm must adapt accordingly. Its behavior begins to change into that of \drrrt\ and \rrt, depending on the sampling success rate for each individual seed. At $d=5$ and $d=6$, \hasrrt's chosen sampling distribution is the whole environment on average $51.6\%$ and $93.7\%$ percent of the time, respectively. This shows that even in extreme cases, \hasrrt\ behaves approximately the same way \rrt\ would. 

For three seeds in the $d=5$ experiment, the runtime is much higher than the remainder of the experiments. This is because while \hasrrt\ readily assumes that skeleton edges can be unreliable, it is more sensitive to skeleton vertices being unreliable. For these three seeds, \hasrrt\ had difficulty finding a valid configuration at a region constructed at one of the skeleton's vertices. 



\section{Conclusion}
\label{sec:conclusion}

In this work we introduce \hasrrtlong\ (\hasrrt), which leverages guidance from a workspace skeleton to efficiently guide the \rrt's expansion process. 
By strategically prioritizing paths available within the workspace, \hasrrt\ can find comparable-cost paths faster than similar methods. 
We also perform an in-depth analysis on the performance of \hasrrt\ with varying qualities of guidance, showing that the method is robust to and can perform efficiently with both minor and major inaccuracies with the provided guidance. 
Our experimental findings underscore the value of incorporating workspace information into motion planning problems where relevant.

\bibliographystyle{styles/IEEEtran}
\bibliography{robotics, biochemistry, geom}
\end{document}